\documentclass{article}
\usepackage{spconf,amsmath,graphicx,booktabs}
\usepackage{amssymb}
\usepackage{hyperref}

\title{Improving transducer-based spoken language understanding with self-conditioned CTC and knowledge transfer}
\name{Vishal Sunder and Eric Fosler-Lussier}
\address{The Ohio State University}
\begin{document}
%\ninept
%
\maketitle
\begin{abstract}
In this paper, we propose to improve end-to-end (E2E) spoken language understand (SLU) in an RNN transducer model (RNN-T) by incorporating a joint self-conditioned CTC automatic speech recognition (ASR) objective. Our proposed model is akin to an E2E differentiable cascaded model which performs ASR and SLU sequentially and we ensure that the SLU task is conditioned on the ASR task by having CTC self conditioning. This novel joint modeling of ASR and SLU improves SLU performance significantly over just using SLU optimization. We further improve the performance by aligning the acoustic embeddings of this model with the semantically richer BERT model. Our proposed knowledge transfer strategy makes use of a bag-of-entity prediction layer on the aligned embeddings and the output of this is used to condition the RNN-T based SLU decoding. These techniques show significant improvement over several strong baselines and can perform at par with large models like Whisper with significantly fewer parameters.
\end{abstract}
\section{Introduction}
\label{sec:intro}
End-to-end spoken language understanding (E2E SLU) is the task of extracting semantic entities from a spoken utterance's acoustic signal instead of first transcribing it. While the latter, cascaded modeling, has been the traditional approach, E2E SLU has become very popular recently, primarily due to its abilities to bypass the harmful cascading effects of ASR errors and of using a smaller sized model. While large language models (LLMs) have become extremely powerful language understanding agents in the recent past, they are not robust to ASR errors \cite{he2023can}. Therefore, for SLU, building-in ASR robustness into models has been an area of active research, with E2E modeling being one of the solutions.

In this paper, we focus on RNN transducer (RNN-T) based E2E SLU. RNN-T was proposed as a recurrent neural network (RNN) based autoregressive ASR model \cite{graves2012sequence}. With the development of sequence encoding models like transformers and conformers \cite{vaswani2017attention, gulati2020conformer}, the RNN is usually switched with these more powerful speech encoders. However, we still refer to these models as RNN-Ts in this paper.

Compared to attention based encoder-decoder models (AED) for ASR \cite{chan2016listen}, RNN-T has the added benefit of being able to process streaming speech while being autoregressive and thus having an implicit language modeling capacity. This makes it a popular choice for ASR modeling in many industrial applications, especially long-form ASR. Thus, many large-scale ASR models are RNN-T based and SLU adaptation is done by using these as seed models. Our proposed approaches in this paper, thus use RNN-T as a base speech processing system. However, it is worth noting that these approaches can be extended to AED based models as well.

In this work, we propose a joint ASR-SLU modeling approach for improving downstream SLU performance. The SLU task is modelled using an RNN-T objective conditioned on the ASR output of the model, which in turn is modelled as a CTC objective \cite{graves2006connectionist}. To make this model E2E differentiable and single-pass during inference, we use the recently proposed self-conditioned version of the CTC loss \cite{nozaki2021relaxing}. Instead of conditioning the RNN-T based SLU loss on the ASR decoding of the model, we condition it on the CTC alignment emission probabilities which serves as a soft decode. We show that this approximation works well in practice. The formulation of this approach is presented in section \ref{sec:prob_form}.

Recent advances in E2E SLU have shown that knowledge can be transferred from LLMs into speech encoders in a fine-grained manner which leads to semantically richer speech representations which in turn improves SLU performance \cite{sunder2023fine}. In this work, we further improve SLU performance by incorporating this knowledge transfer (KT) technique and propose a novel extension to this idea by incorporating an auxiliary prediction of bag-of-entities present in an utterance.

In particular, we propose to incorporate a bag-of-entities prediction layer into our model. The soft prediction from this layer is added into the RNN-T joint network which acts as a soft prior on the prediction. This helps the RNN-T prediction network perform an informed decoding of the slots and values thus improving performance. A detailed and formal description of this is provided in section \ref{subsec:KT}.

We highlight the following novel contributions in this work. First, we propose a fully E2E and differentiable cascading of ASR and SLU by incorporating self-conditioned CTC based ASR objective into a transducer based SLU model. While this has been explored for ASR and speech translation tasks, we are the first to introduce this into an E2E SLU model to the the best of our knowledge. Second, we improve previously proposed LLM-based knowledge transfer techniques \cite{sunder2023fine, sunder2022tokenwise} by incorporating conditioning of the transducer decoder over bag-of-entities prediction.

In the next section, we provide a literature review on the topic. Section \ref{sec:prob_form} is the formulation of the problem and details our approach to it. Next, in section \ref{sec:exp_res} we go through our implementation details, present our results and provide a thorough quantitative and qualitative analysis of our findings.

\section{Related work}
\label{sec:rel_work}
\textbf{Cascaded ASR-NLU}: Most research in ASR and natural language understanding (NLU) has advanced in a somewhat disjoint manner. While NLU models have made remarkable progress over the years, they still greatly underperform in the presence of ASR errors \cite{kim2021robust, faruqui2022revisiting, he2023can}. Thus, a lot of work has focused on building interfaces between ASR and NLU. Along these lines, Raju et al. \cite{raju2021joint} explore ways of stitching together ASR and NLU systems and Seo et al. \cite{seo2022integration} build an interface for E2E training of large speech and language models. Arora et al. \cite{arora2022token} build an E2E ASR-NLU model where the NLU submodel is connected with the ASR submodel by a cross-attention mechanism.

% Namazifar et al. \cite{namazifar2021warped} introduced the notion of "warped" language models where masked language models like BERT were pretrained with simulated ASR errors, hence becoming more ASR robust during NLU adaptation. Another approach proposed by Ruan et al. \cite{ruan2020towards} used pseudo-ASR transcripts for data augmentation. A similar idea was proposed by Sunder et al. \cite{sunder2022building} where they augment their NLU training data with errorful text using a neural ASR error generator.

% Another popular paradigm for improving ASR robustness in NLU systems is building interfaces between ASR and NLU. Along these lines, Raju et al. \cite{raju2021joint} explore ways of stitching together ASR and NLU systems and Seo et al. \cite{seo2022integration} build an interface for E2E training of large speech and language models. Arora et al. \cite{arora2022token} build an E2E ASR-NLU model where the NLU submodel is connected with the ASR submodel by a cross-attention mechanism.

\textbf{E2E SLU}: Over the years, perhaps the most popular design choice for E2E SLU has been to pretrain an encoder-decoder based ASR model and then adapt this model for SLU in an E2E manner. Kuo et al. \cite{kuo2020end} train an E2E attention based model and study how these models can be adapted to new SLU domains without any ASR domain adaptation. RNN-T based E2E SLU models have similarly been developed \cite{thomas2021rnn, raju2021end}. Recently, scaling these models with large amounts of data for ASR pretraining and SLU adaptation has shown promise for the task \cite{arora2023universlu, huang2023leveraging}. Self-supervised speech models without ASR adaptation have also shown a great capability to perform E2E SLU \cite{wang2021fine}.

% Decoding a sequence of semantic entities from speech has also garnered a lot of attention primarily since encoder-decoder based ASR models can be effectively adapted to generate a sequence of semantic entities. Some initial studies in this regard were conducted in \cite{haghani2018audio, serdyuk2018towards} where various design choices were explored for effective E2E SLU.  

\textbf{Auxiliary CTC for speech tasks}: The connectionist temporal classification (CTC) \cite{graves2006connectionist} is a non-autoregressive sequence tagging objective primarily used for E2E ASR. One of the many benefits of CTC is that it does not need additional parameters to be trained except a classification layer. Therefore, for speech tasks, having an auxiliary CTC objective can help speech align with the corresponding transcript without needing too many excessive parameters. This advantage was utilized to improve the attention-based ASR model by Kim et al. \cite{kim2017joint}. A similar use of the CTC objective \cite{yan2022ctc} shows its advantage in text-to-text and speech-to-text translation. Furthermore, recent work has also shown the effectiveness of CTC objectives for massive multilingual ASR by having an intermediate CTC predict the language ID \cite{chen2023improving}.

For SLU, Wang et al. \cite{wang2023end} show that auxiliary CTC improves intent recognition performance in pretrained self-supervised models. However, their model does not use self-conditioning on intermediate CTC emissions.

\textbf{Knowledge transfer for speech tasks}: With the advent of large language models, techniques have emerged to transfer their expertise gained through large scale training into speech models. Work by Sunder et al. \cite{sunder2022towards, sunder2022tokenwise, sunder2023fine, sunder2023convkt} show how embedding-level alignment can be achieved between speech and text using contrastive learning. Huang et al. \cite{huang2020leveraging} also proposed an alignment strategy between speech and BERT embeddings for better speech understanding. Even ASR performance has shown significant improvements with similar knowledge transfer strategies \cite{kubo2022knowledge, sunder2023fine}.

\section{Problem formulation}
\label{sec:prob_form}
In the below formulation, we use the following notations: Let the speech be represented by $X = [x_t \in \mathbb{R}^d | t = 1, .., T_X]$ where $x_t$ is a speech frame at time $t$. Next, let the sequence of SLU tags be $S = [s_i \in \mathcal{V}_S| i = 1, .., T_S]$ where $s_i$ is an SLU tag from the set of SLU tag vocabulary $\mathcal{V}_S$. Similarly, the transcript of $X$ is $Y = [y_i \in \mathcal{V}_Y| i = 1, .., T_Y]$ where $y_i$ is the token from the set of vocabulary $\mathcal{V}_Y$.

\subsection{Background on RNN transducers}
\label{subsec:rnnt}
RNN transducers are autoregressive speech-to-text models that learn an alignment of length $T_X + T_Y$ between speech and text. When used for E2E SLU, they learn an alignment of length $T_X + T_S$ between speech and SLU tags. In general, given alignment $A$, RNN-T estimates $P(Y|X)$ as,
\begin{align*}
    \begin{split}
        P(Y|X) &\approx \sum_{A \in \mathcal{B}_{rnnt}^{-1}(Y)} P(A|X) \\
        &\approx \sum_{A \in \mathcal{B}_{rnnt}^{-1}(Y)} \prod_{t = 1}^{T_X + T_Y} P(a_t | y_{1..t'}, X)
    \end{split}
\end{align*}
Here, $\mathcal{B}_{rnnt}(A)$ is a collapsing function that converts $A$ to $Y$, $\mathcal{B}_{rnnt}^{-1}(Y)$ is the corresponding one-to-many mapping from $Y$ to $A$ and $y_{1..t'} = \mathcal{B}_{rnnt}(a_{1..t-1})$.  The RNN-T uses a transcription network to encode the speech $X$, a prediction network to encode the token sequence $y_{1..t'}$ and a joint network that combines the two to estimate $P(Y|X)$.

\subsection{Background on Self-Conditioned CTC}
\label{subsec:scctc}
Self-conditioned CTC \cite{nozaki2021relaxing} was proposed to improve CTC-based ASR models by having intermediate CTC objectives in neural network layers and adding the prediction of these intermediate layers into higher layers. The model thus performs iterative prediction refinements across layers and this relaxes the conditional independence assumption among predicted tokens in CTC. More formally, for a CTC alignment $A$ and a collapsing function $\mathcal{B}_{ctc}$, CTC estimates,
\begin{align*}
    \begin{split}
        P(Y|X) &\approx \sum_{A \in \mathcal{B}_{ctc}^{-1}(Y)} P(A|X) \\
        &\approx \sum_{A \in \mathcal{B}_{ctc}^{-1}(Y)} \prod_{t = 1}^{T_X} P(a_t | X)
    \end{split}
\end{align*}
Notice that unlike RNN-T, CTC is non-autoregressive. To relax this, SCTC makes the following change,
\begin{align*}
% \label{eqn:scctc}
    \begin{split}
        &P(Y|X) \approx \prod_{i=1}^{K} \sum_{A_i \in \mathcal{B}_{ctc}^{-1}(Y)} \prod_{t = 1}^{T_X} P(a_t^i | X, Z_{i-1}) \\
        &Z_i = \text{Linear}_i^1(\underbrace{\text{Softmax}(\text{Linear}_i^2(\text{Conformer}_i(X_{i-1}+Z_{i-1})))}_{\text{Emissions from the $i^{th}$ CTC layer}})
    \end{split}
\end{align*}
Here, we assume that there are $K$ intermediate CTC layers and the emissions from each is used to compute the emissions from the next. $\text{Conformer}_i$ refers to a sequence of conformer layers upto the $i^{th}$ CTC, $X_i$ is the conformer output where the $i^{th}$ CTC is present and $Z_0 = \textbf{0}$.

\subsection{Proposed joint ASR and SLU modeling}
\label{subsec:joint}

\begin{figure}
    \hfill
    \centering
    \centerline{\includegraphics[width=\columnwidth]{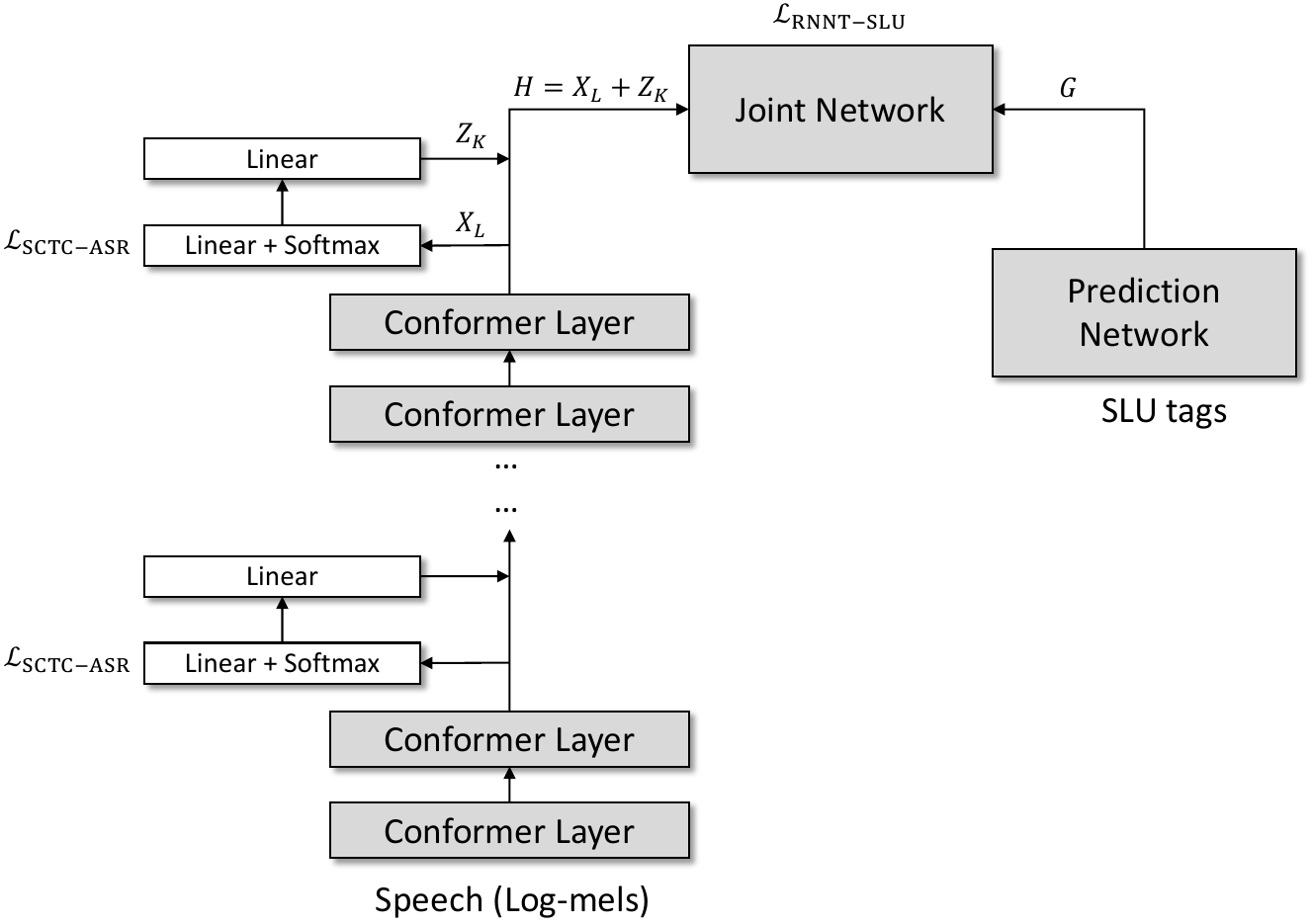}}
\caption{A sequence of $L$ conformer layers serve as the transcription network. After every other conformer layer, we have an intermediate CTC loss with self-connections.} 
\label{fig:model_sctc}
\end{figure}

The E2E SLU task consists of predicting a sequence of entities from the acoustics of a spoken utterance. We model the sequence of entities as a sequence of slots and values. For example, in the utterance, "how cold is it outside today", from the SLURP dataset, the SLU tag is represented as: \texttt{IN-weather\_query} \textit{cold} \texttt{b-weather\_descriptor} \textit{today} \texttt{b-date}. Here, the first token is always the intent.

Let $A_Y$ be a valid CTC alignment between $Y$ and $X$, and $\mathcal{B}_{ctc}$ be the collapsing function that converts $A_Y$ to $Y$. In the same way, $A_S$ is a valid RNN-T alignment between $S$ and $X$ and $\mathcal{B}_{rnnt}$ is the corresponding collapsing function.

%  In this work, we train a model to maximize the likelihood of the joint distribution, $P(Y, S | X)$. This is modelled as,
% \begin{align*}
%     \begin{split}
%         P(Y, S | X) &= P(S | Y, X)P(Y | X) \\
%         &\approx \sum_{A_Y \in \mathcal{B}^{-1}_{ctc}} \sum_{A_S \in \mathcal{B}^{-1}_{rnnt}} P(A_S | A_Y, X) P(A_Y | X) \\
%         &\approx \underbrace{\sum_{A_S \in \mathcal{B}^{-1}_{rnnt}} P(A_S | Z, X)}_{P_{\text{rnnt}}^{\text{slu}}} \underbrace{\sum_{A_Y \in \mathcal{B}^{-1}_{ctc}} P(A_Y | X)}_{P_{\text{ctc}}^{\text{asr}}} \\
%     \end{split}
% \end{align*}
% Here, $Z = \text{Linear}(\text{Softmax}(\text{Linear}(\text{Conformer(X)})))$ is the SC-CTC's last layer emissions. The RNN-T portion of the above equation predicts the SLU tags conditioned on the CTC's soft prediction which in-turn is trained to predict the ASR transcription. Note that the CTC portion is modelled as a SC-CTC as shown in. and the output of the last SC-CTC layer is used by the transducer. The loss function can be defined as,
% \begin{align}
%     \begin{split}
%         \mathcal{L}_{\text{joint}} &= -\lambda \text{log}(P_{\text{rnnt}}^{\text{slu}}) - (1 - \lambda) \text{log}(P_{\text{ctc}}^{\text{asr}}) \\
%         &= \lambda \mathcal{L}^{\text{slu}}_{\text{rnnt}} + (1 - \lambda) \mathcal{L}^{\text{asr}}_{\text{ctc}}
%     \end{split}
% \end{align}

In this work, we train a model to maximize the likelihood of the joint distribution, $P(Y, S | X)$. This is modelled as,
\begin{align*}
    \begin{split}
        P(Y, S &| X) = P(S | Y, X)P(Y | X) \\
        &\approx \sum_{A_S \in \mathcal{B}^{-1}_{rnnt}(S)} \sum_{A_Y \in \mathcal{B}^{-1}_{ctc}(Y)} P(A_S | A_Y, X) P(A_Y | X) \\
    \end{split}
\end{align*}

The above likelihood is difficult to compute during training and would require a two-pass decoding strategy during inference. Instead, we make the above likelihood tractable and one-pass by using the SCTC model. Following from section \ref{subsec:scctc}, we rewrite the above likelihood as,
% \small
\begin{align*}
    \begin{split}
        P(Y, S | X) &\approx \underbrace{\sum_{A_S} P(A_S | Z_K, X)}_{P_{\text{RNNT}-\text{SLU}}} \underbrace{\prod_{i=1}^{K} \sum_{A_Y^i} P(A_Y^i | X, Z_{i-1}}_{P_{\text{SCTC}-\text{ASR}}})
    \end{split}
\end{align*}
% \normalsize
In $P_{\text{RNNT}-\text{SLU}}$, $Z_K$ is the emission from the last CTC layer and $Z_i$ is defined in section \ref{subsec:scctc}. This serves as a soft prediction and a proxy for $A_Y$. $Z_K$ is simply added to the last conformer output before going into the joint network, thus the final output of the transcription network is $X_L + Z_K$. $P_{\text{SCTC}-\text{ASR}}$ is exactly the SCTC likelihood defined in section \ref{subsec:scctc}. The joint loss function can be defined as,
\begin{align}
\label{eqn:l_jnt}
    \begin{split}
        \mathcal{L}_{\text{JNT}} &= -\lambda \text{log}(P_{\text{RNNT}-\text{SLU}}) - (1 - \lambda) \text{log}(P_{\text{SCTC}-\text{ASR}}) \\
        &= \lambda \mathcal{L}_{\text{RNNT}-\text{SLU}} + (1 - \lambda) \mathcal{L}_{\text{SCTC}-\text{ASR}}
    \end{split}
\end{align}

An overview of the model is shown in figure \ref{fig:model_sctc}. Let the output from the $L$ layer transcription network be $H = X_L + Z_K$ which is a sequence of vectors, $H = [h_t \in \mathbb{R}^{768}| t = 1,.., T_X]$. The output from the prediction network, modeled as a single layer LSTM, is $G = [g_u \in \mathbb{R}^{1024}| u = 1,.., T_S]$. Then, $P_{\text{RNNT}-\text{SLU}}$ is computed using dynamic programming from the emissions, $P(.|h_t, g_u)$, from the RNN-T which is modelled using a joint network as,
\begin{align*}
    \begin{split}
        P(.|h_t, g_u) = \text{Softmax}(W_{out}\text{tanh}(W_{enc}h_t + W_{pred}g_u + b))
    \end{split}
\end{align*}
Here, $W_{out} \in \mathbb{R}^{|\mathcal{V}_S|\times 256}$, $W_{enc} \in \mathbb{R}^{256\times 768}$, $W_{pred} \in \mathbb{R}^{256\times 1024}$ and $b \in \mathbb{R}^{256}$ are learnable parameters.

\subsection{Proposed knowledge transfer}
\label{subsec:KT}
We propose to further improve the above model by transferring knowledge from a large language model. In this paper, we employ BERT\footnote{\texttt{https://huggingface.co/bert-base-uncased}}, but our method is applicable to any LLM. For this, we utilize a method proposed by Sunder et al. \cite{sunder2023fine} and extend it further for improved performance.

Our method comprises of two stages. In the first stage, an ASR pretraining is conducted with knowlege transfer (KT) from BERT. We adapt this model for SLU in the second stage using a novel bag-of-entities prediction loss.\\

\begin{figure}
    \hfill
    \centering
    \centerline{\includegraphics[width=\columnwidth]{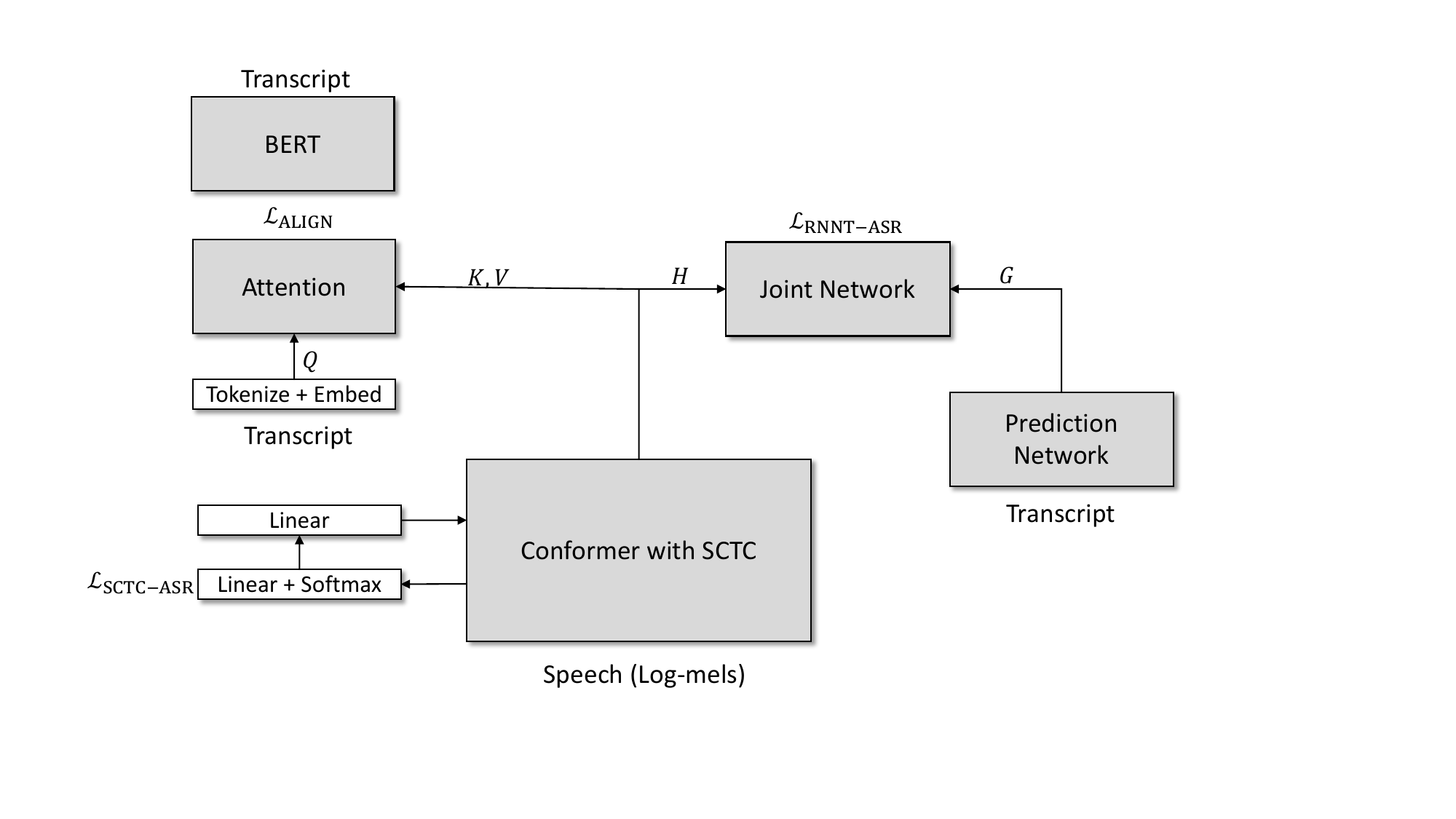}}
\caption{ASR pretraining with knowledge transfer. The transcription network is the same as figure \ref{fig:model_sctc}. Both RNN-T and SCTC losses are computed against the transcription.} 
\label{fig:model_kt_asr}
\end{figure}

\noindent \textbf{ASR Pretraining with KT (figure \ref{fig:model_kt_asr})}: In this stage, we seek to transfer the knowledge from BERT into the speech encoder. The transcript $Y$ is fed to BERT to generate a sequence of embeddings $B_Y \in \mathbb{R}^{T_Y^{'} \times 768}$ which acts as the teacher signal. A student signal, $B_X \in \mathbb{R}^{T_Y^{'} \times 768}$, is generated from the speech encoder using an attention mechanism as,
\begin{align*}
    \begin{split}
        B_X = \text{Attention}(\text{Embedding}(\text{Tokenize}(Y)), H)
    \end{split}
\end{align*}
Here, $\text{Embedding}(\text{Tokenize}(.))$ operation is the same as BERT. $B_X$ is now aligned with $B_Y$ using a contrastive loss,

\footnotesize
\begin{align*}
    \begin{split}
        \mathcal{L}_{\text{ALIGN}} = -\frac{\tau}{2b}\sum_{i=1}^{b}(&\log\frac{\exp(s_{ii}/\tau)}{\sum_{j=1}^{b} \exp(s_{ij}/\tau)} + \log\frac{\exp(s_{ii}/\tau)}{\sum_{j=1}^{b}\exp(s_{ji}/\tau)})
    \end{split}
\end{align*}
\normalsize
$B_Y$ and $B_X$ across a batch are row-wise concatenated such that $B_Y$ and $B_X$ are now $\in \mathbb{R}^{b \times 768}$, where $b$ is the sum of all sequence lengths in a batch and $s_{ij}$ refers to the cosine similarity between the $i^{th}$ and $j^{th}$ rows of $B_Y$ and $B_X$. The final pretraining loss is defined as,
\begin{align}
\label{eqn:l_pt}
    \begin{split}
        \mathcal{L}_{\text{ASR}-\text{KT}} = \lambda \mathcal{L}_{\text{RNNT}-\text{ASR}} + (1 - \lambda) \mathcal{L}_{\text{SCTC}-\text{ASR}} + \alpha \mathcal{L}_{\text{ALIGN}}
    \end{split}
\end{align}
During pretraining, RNN-T loss is computed against the transcriptions, not the SLU tags. We set $\lambda = 0.5$, $\alpha = 1.0$ and $\tau = 0.07$. Note that the SCTC component is also used here.\\

\begin{figure}
    \hfill
    \centering
    \centerline{\includegraphics[width=\columnwidth]{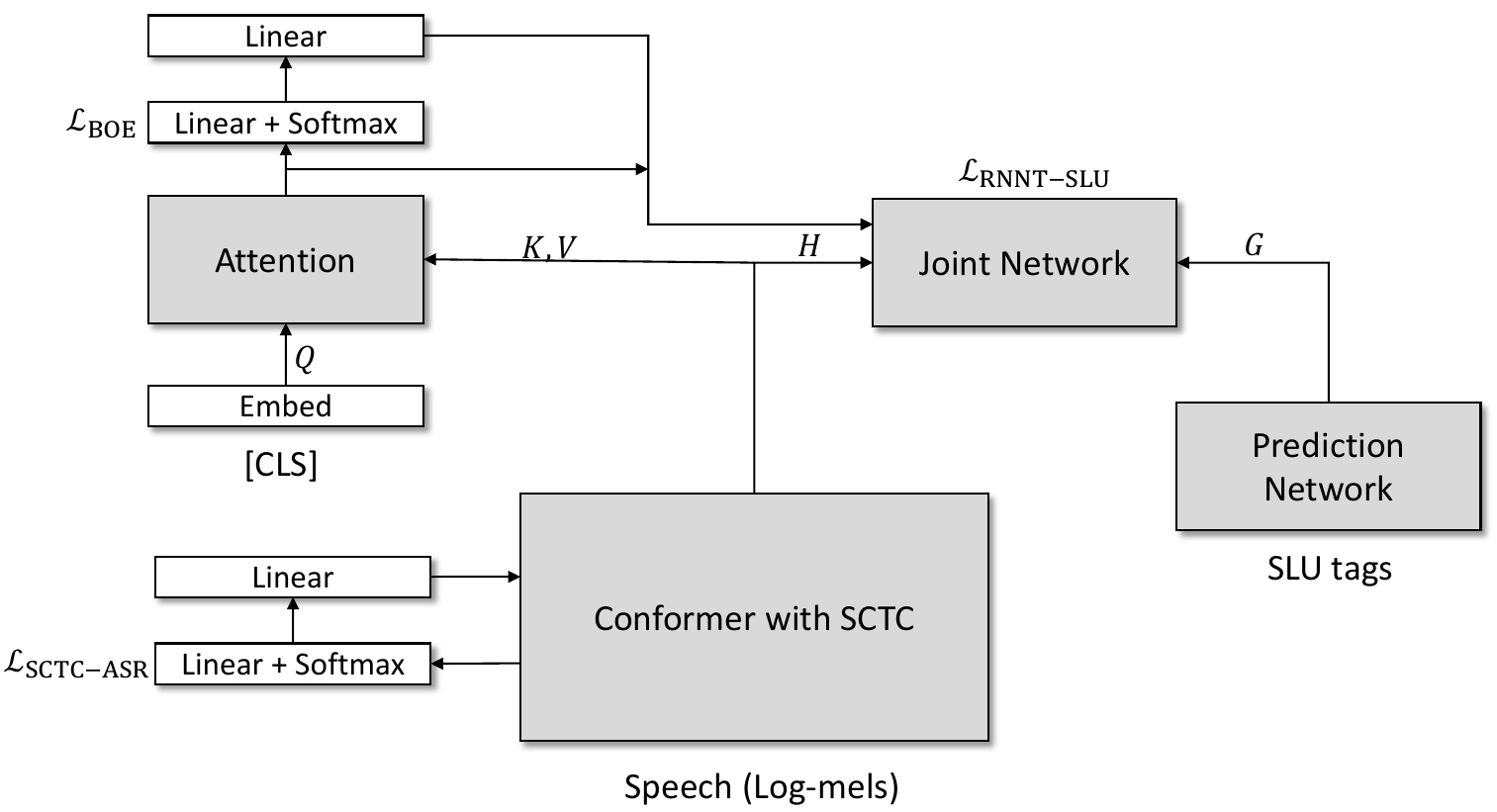}}
\caption{SLU adaptation with knowledge transfer. After the ASR pretraining step in figure \ref{fig:model_kt_asr}, the model is adapted for SLU where the utterance level representation from the Attention block is utilized for predicting the bag-of-entities and is added to the joint network as in equation.} 
\label{fig:model_kt_slu}
\end{figure}

\noindent \textbf{SLU adaptation with KT (figure \ref{fig:model_kt_slu})}: Pretraining with KT as explained above makes the conformer encoder semantically rich. Furthermore, the attention layer can produce embeddings which are close to BERT embeddings. To make use of the attention layer during inference, we utilize the [CLS] embedding as the attention query during SLU adaptation. The [CLS] embedding is an utterance level representation which is close to BERT's [CLS] embedding. In the acoustic domain, we call this $x_{\text{[CLS]}}$, i.e. $x_{\text{[CLS]}} = \text{Attention}(\text{Embedding}(\text{[CLS]}), H)$.

Similar to Sunder et al. \cite{sunder2023fine}, we integrate $x_{\text{[CLS]}}$ into the joint network of the RNN-T model. In addition, we propose a simple yet novel approach of providing a soft signal to the RNN-T decoder about the slot mentions in the SLU tag. For example, the SLU tag, \texttt{IN-weather\_query} \textit{cold} \texttt{b-weather\_descriptor} \textit{today} \texttt{b-date}, has the following slots: [\texttt{IN-weather\_query}, \texttt{weather\_descriptor}, \texttt{date}]. We have an auxiliary classification layer after the attention layer which predicts this set as a multi-hot, bag-of-entities (BOE) vector, $Y_{\text{BOE}}$, as shown below,
\begin{align*}
    \begin{split}
        P_{\text{BOE}}(.|X) &= \text{Softmax}(\text{Linear}(x_\text{[CLS]})) \\
        \mathcal{L}_{\text{BOE}} &= -\text{log}(P_{\text{BOE}}(Y_{\text{BOE}}|X))
    \end{split}
\end{align*}
Here, $Y_{\text{BOE}}$ is the ground truth, L-1 normalized representation of the multi-hot representation described above.

Finally, we update the joint network equation as follows,
\small
\begin{align}
\label{eqn:lambda}
    \begin{split}
        &\gamma = \sigma(W_h h_t + W_g g_u + b') \odot (W_b P_{\text{BOE}}(.|X) + W_c x_{\text{[CLS]}}) \\
        &P(.|h_t, g_u) = \text{Softmax}(W_{out}\text{tanh}(W_{enc}h_t + W_{pred}g_u + b + \gamma))
    \end{split}
\end{align}
\normalsize
Here, $\gamma$ is the auxiliary information from the predicted bag-of-entities and the utterance-level representation, $x_{\text{[CLS]}}$. The sigmoid gate, $\sigma(.)$ controls this information as a function of the current position ($t,u$) in the RNN-T trellis. The RNN-T loss is now computed from the above emission probability. We call this loss $\mathcal{L}_{\text{RNNT}-\text{SLU}}^{\text{CLS, BOE}}$.

The loss function of the SLU adaptation is given as,
\begin{align}
\label{eqn:l_jnt_kt}
    \begin{split}
        \mathcal{L}_{\text{JNT}-\text{KT}} = \lambda \mathcal{L}_{\text{RNNT}-\text{SLU}}^{\text{CLS, BOE}} + (1 - \lambda) \mathcal{L}_{\text{SCTC}-\text{ASR}} + \beta \mathcal{L}_{\text{BOE}}
    \end{split}
\end{align}
Here, $\lambda = 0.5$ and $\beta = 0.1$. 

% An illustration of this model is given in figure \ref{fig:model_kt_slu}.

\section{Experiments and results}
\label{sec:exp_res}
\begin{table*}
    \centering
    \resizebox{1.4\columnwidth}{!}{
    \begin{tabular}{@{}lcccc@{}}\toprule
        Model & Precision & Recall & SLU-F1 & Intent-Acc\\
        \midrule
        \textbf{Previous models} &&&& \\
        \midrule
        CTI \cite{seo2022integration} & - & - & 74.66 & 86.92 \\
        Sunder et al. \cite{sunder2023fine} & - & - & 76.96 & 87.95 \\
        Branchformer \cite{peng2022branchformer} & - & - & 77.70 & 88.10 \\
        Arora et al. \cite{arora2023integrating} & - & - & 78.50 & -  \\
        CIF \cite{dong2023cif} & - & - & 78.67 & 89.60 \\
        HuBERT-Large$^\dagger$ \cite{wang2021fine} & 80.54 & 77.44 & 78.96 & 89.37 \\
        UniverSLU \cite{arora2023universlu} & - & - & 79.50 & -  \\
        Whisper$^\dagger$ \cite{radford2023robust} & - & - & 79.70 & - \\
        NeMo-Large$^\dagger$ \cite{huang2020leveraging} & \textit{84.31} & \textit{80.33} & \textit{82.27} & \textit{90.14} \\
        \midrule
        \textbf{Our baseline models} &&&& \\
        \midrule
        (1) ASR finetune RNN-T $\rightarrow$ SLU adapt RNN-T & 79.22 & 71.24 & 75.02 & 85.92 \\
        (2) ASR finetune RNN-T w/ $\mathcal{L}_{\text{ALIGN}}$ $\rightarrow$ SLU adapt w/ $\mathcal{L}_{\text{RNNT}-\text{SLU}}^{\text{CLS}}$ & 79.88 & 73.95 & 76.80 & 87.95 \\
        \midrule
        \textbf{Our proposed models (Joint ASR+SLU)} &&&& \\
        \midrule
        (3) No ASR finetuning $\rightarrow$ $\mathcal{L}_{\text{JNT}}$(\ref{eqn:l_jnt}) & \textbf{82.83} & 74.91 & 78.67 & 88.00\\
        (4) ASR finetune w/ RNN-T + SCTC $\rightarrow$ $\mathcal{L}_{\text{JNT}}$(\ref{eqn:l_jnt}) & 82.01 & 75.50 & 78.62 & 87.89 \\
        \midrule
        \textbf{Our proposed models (Knowledge transfer)} &&&& \\
        \midrule
        (5) $\mathcal{L}_{\text{ASR}-\text{KT}}$(\ref{eqn:l_pt}) $\rightarrow$ $\mathcal{L}_{\text{JNT}}$(\ref{eqn:l_jnt}) & 82.07 & 74.71 & 78.22 & 87.90 \\
        (6) $\mathcal{L}_{\text{ASR}-\text{KT}}$(\ref{eqn:l_pt}) $\rightarrow$ $\mathcal{L}_{\text{JNT}-\text{KT}}$ w/o BOE component & 81.65 & 76.67 & 79.02 & 89.20 \\
        (7) $\mathcal{L}_{\text{ASR}-\text{KT}}$(\ref{eqn:l_pt}) $\rightarrow$ $\mathcal{L}_{\text{JNT}-\text{KT}}$(\ref{eqn:l_jnt_kt}) & 82.12 & \textbf{77.22} & \textbf{79.59} & \textbf{89.68} \\
         \bottomrule
    \end{tabular}}
    \caption{Our proposed models are first pretrained for ASR on the Fisher 2000 hour dataset using $(\mathcal{L}_{\text{SCTC}-\text{ASR}}, \mathcal{L}_{\text{RNNT}-\text{ASR}})$. Then, these models are ASR finetuned on SLURP $\xrightarrow[\text{by}]{\text{followed}}$ SLU adapted on SLURP. This is indicated by $(...)$ $\rightarrow$ $(...)$ using different combinations of the proposed losses. $^\dagger$ indicates that these models either use significantly more pretraining data or significantly more parameters or both compared to our proposed models. All our models have $\leq$ 100 million parameters.} 
    \label{tab:results_main}
\end{table*}

\begin{table}
    \centering
    \resizebox{\columnwidth}{!}{
    \begin{tabular}{@{}ccccccc@{}}\toprule
    SCTC-1 & SCTC-2 & SCTC-3 & Precision & Recall & SLU-F1 & Intent-Acc\\
    \midrule
    ASR & ASR & ASR & \textbf{82.83} & \textbf{74.91} & \textbf{78.67} & \textbf{88.00} \\
    \midrule
    ASR & ASR & SLU & 76.50 & 74.58 & 76.03 & 86.90 \\
    ASR & SLU & SLU & 76.10 & 74.32 & 75.75 & 87.00 \\
    SLU & SLU & SLU & 75.75 & 74.67 & 75.21 & 86.60 \\

    \bottomrule
    \end{tabular}}
    \caption{Effect of using ASR/SCTC objectives on the three intermediate SCTC layers of our 6-layer conformer model.}
    \label{tab:results_sctc_asr}
\end{table}

\subsection{Experiment settings}
\label{subsec:exp_sett}
\textbf{Model details}: For our speech encoder, we use a 6-layer conformer with 768 hidden units and 12 attention heads. The prediction network is a 1-layer LSTM with 1024 hidden units. Our tokenization strategy is a simple character-based tokenization for ASR. For SLU, we treat the intent labels and slot labels as additional tokens to our ASR vocabulary, with the slots being tokenized at the character level.

We also utilize SpecAugment \cite{park2019specaugment} and sequence noise injection \cite{saon2019sequence} as regularization strategies. We obtain sequences of 40 dimensional log-mel filterbank (LFB) features from speech resampled at 8kHz. We also add $\Delta$ and $\Delta^2$ coefficients to the LFB features and stack two consecutive frames, skipping every other frame resulting in a feature dimension of 240 which gets fed into the conformer.

\textbf{Data}: As a first step, we train the above model on the 2000-hour Fisher dataset with the ASR objective. This consists of a combination of $\mathcal{L}_{\text{RNNT}-\text{ASR}}$ and $\mathcal{L}_{\text{SCTC}-\text{ASR}}$. Our SLU models were initialized with this pretrained ASR model.

For SLU, we perform experiments on the SLURP \cite{bastianelli2020slurp}, which is a speech dataset with both ASR and SLU labels. It contains 84 hours of speech data in the training set and 6.9 hours and 10.2 hours of data in the dev and test set respectively. For evaluation, we use the metrics proposed in \cite{bastianelli2020slurp}.

\subsection{Discussion on results}
\label{subsec:res}
The main results of our experiments are shown in table \ref{tab:results_main}. We also include the latest published results on the SLURP SLU task for comparison. Rows (1) and (2) are the baselines without the proposed joint ASR, SLU setup. Note that all results that follow (rows (3) to (7)) give better SLU performance. All the proposed models are pretrained for ASR on the Fisher 2000 hour dataset using $\mathcal{L}_{\text{SCTC}-\text{ASR}} + \mathcal{L}_{\text{RNNT}-\text{ASR}}$.

\textbf{Effect of SCTC}: Rows (3) and (4) show the result of including an auxiliary SCTC-based ASR loss for RNNT-based SLU as formulated in section \ref{subsec:joint}. We note a significant improvement in slot-filling performance. We note that even without an ASR finetuning step on SLURP, directly adapting the model for SLU gives an improved performance (row (3)). This may be because the SCTC-based ASR objective takes care of the acoustic domain adaptation and "informing" the decoder about the entities present. However, we also notice that the intent prediction accuracy does not undergo a significant change. Since good intent prediction depends on a good utterance-level representation, we use our proposed integration of the utterance-level $x_{\text{[CLS]}}$ vector for the same.

\textbf{Analysis of the proposed KT mechanisms}: Rows (5), (6) and (7) show the results of the proposed KT-based ASR pretraining and SLU adaptation in section \ref{subsec:KT}. In row (5), we finetune the model for ASR using $\mathcal{L}_{\text{ASR}-\text{KT}}$ defined in equation \ref{eqn:l_pt} and adapt it for SLU using $\mathcal{L}_{\text{JNT}}$ (equation \ref{eqn:l_jnt}). We do not observe an improvement in performance from the KT based pretraining. However, when we ASR-finetune the model with $\mathcal{L}_{\text{ASR}-\text{KT}}$ and adapt it for SLU by incorporating KT into the SLU framework, as shown in row (6), we see improvements across the board. We note that as $x_\text{[CLS]}$ is now semantically rich with the KT based pretraining, it leads to better entity extraction and intent prediction. Thus, even without the BOE component, the performance improves, as shown in row (6). 

Row (7) shows our best performing model when ASR-fintuned using $\mathcal{L}_{\text{ASR}-\text{KT}}$ (equation \ref{eqn:l_pt}) and SLU-adapted using $\mathcal{L}_{\text{JNT}-\text{KT}}$ (equation \ref{eqn:l_jnt_kt}). The prediction ($\mathcal{L}_{\text{BOE}}$) and integration (equation \ref{eqn:lambda}) of the bag-of-entities information serves as an effective prior over the intent and slots present in the utterance.

From rows (5) and (6), we see that the KT incorporation improves recall but degrades precision. Having the BOE component recovers the precision by avoiding the false positives in a better way and serving as an effective regularization.

% makes the model extract more entities possibly due to having an overconfident estimate of the utterance semantics. As a result, false negatives are reduced (improved recall) but false positives are increased (worse precision). Having the BOE component in row (7) recovers back the precision score by avoiding the false positives in a better way. 
% This suggests that the BOE component serves as an effective regularization.
% and is similar, in principle, to the self-conditioning in the SCTC objectives.

\textbf{Comparing with previous models}: When comparing with previously reported results on SLURP, our model comes close to Whisper's SLU performance and is second only to NeMo-Large. It is worth noting that the above version of Whisper is pretrained on a very large amount of data with the ASR objective and has close to 800 million parameters. NeMo-Large is also pretrained on roughly 25,000 hours of speech. Compared to this, our model utilizes 2,000 hours of pretraining data. Furthermore, the proposed SCTC-based joint training can be easily integrated with the above large industrial-scale models. Also, note that using the proposed techniques, we are able to outperform HuBERT-Large based SLU which is about 3 times larger than our models.

\textbf{An analysis on SCTC objectives}: The proposed joint training in equation \ref{eqn:l_jnt} trained the SCTC layers using the speech transcriptions as targets, i.e. SCTC was an ASR objective. However, we can also train SCTC across various layers of the conformer using SLU tags as targets. Thus, for our 6-layer conformer model with 3 intermediate SCTC layers, we explore using SLU objectives as shown in table \ref{tab:results_sctc_asr}. 

We observe that anytime we introduce an SLU objective in any of the SCTC layers, the performance is degraded. The performance is gradually improved as we incorporate ASR until all SCTC layers are ASR objectives. We hypothesize that the SLU task is difficult to perform non-autoregressively and therefore using SCTC for SLU can give a soft prediction which is noisy and may lead to cascading of errors. Furthermore, the ASR objective in SCTC enables the model, in a fully end-to-end differentiable way, to perform a cascaded ASR-SLU processing of speech where the ASR can help the SLU. The ASR itself benefits from layerwise iterations as in SCTC and this helps autoregressive decoding of the SLU tags.

\begin{figure}
    \hfill
    \centering
    \centerline{\includegraphics[width=0.8\columnwidth]{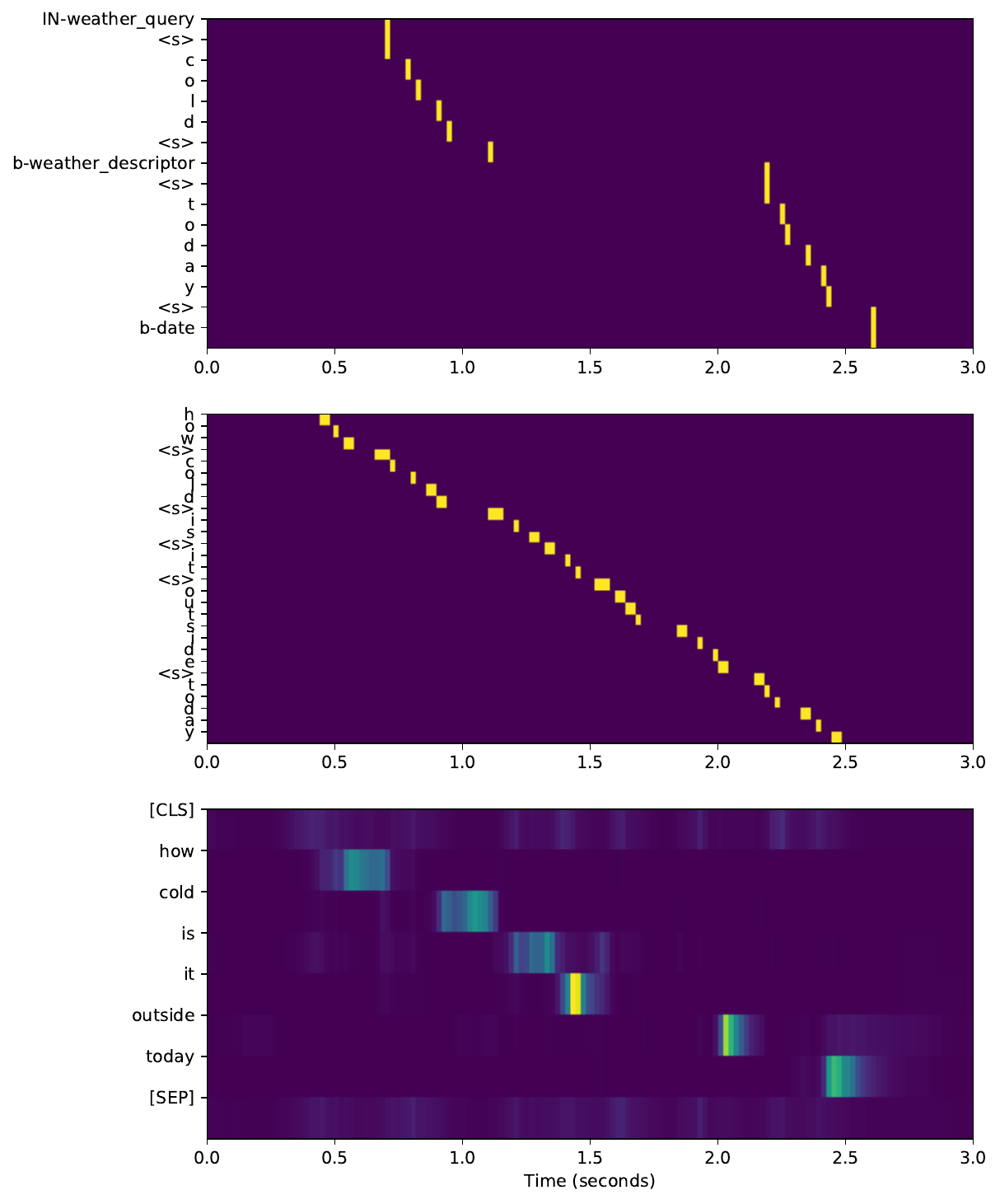}}
\caption{Alignment at different levels for a sample utterance \textit{"how cold is it outside today"}. \textit{Top:} The RNN-T alignment between the SLU tags and the speech, \textit{Middle:} character-level alignment at the last SCTC layer, \textit{Bottom:} subword-level alignment sfrom the attention layer during KT pretraining.} 
\label{fig:alignment}
\end{figure}

\textbf{Alignments learnt by our model}: Our proposed model learns three levels of alignments between the speech input and the utterance semantics. These are shown in figure \ref{fig:alignment}. The first is the alignment between the speech signal and the semantic entities in the speech utterance which is learnt by the RNN-T. We see that the model emits the tokens when the corresponding entity is mentioned. The middle part of the figure shows the alignment obtained by the last CTC layer. The bottom part shows the attention map of the wordpiece tokenization of the utterance and the speech input in the Attention layer used for knowledge transfer (figure \ref{fig:model_kt_asr}).

\section{Conclusion}
\label{sec:con}
This work presented a joint modeling approach for ASR and SLU and showed that this joint modeling can help the SLU task. We explicitly conditioned the output of the RNN-T based SLU model on the output of the ASR model by using a self-conditioned CTC objective. Further improvements in SLU were proposed by using a fine-grained knowledge transfer strategy from BERT embeddings into conformer based acoustic embeddings. Future work should look at how this model scales to large datasets and larger model sizes.

% References should be produced using the bibtex program from suitable
% BiBTeX files (here: strings, refs, manuals). The IEEEbib.bst bibliography
% style file from IEEE produces unsorted bibliography list.
% -------------------------------------------------------------------------
\bibliographystyle{IEEEbib}
\bibliography{strings,refs}

\end{document}